**Detecting Glaucoma Across Multi-ethnic Myopic and Non-Myopic Populations Using an Uncertainty-Aware Vision Transformer: A Multicentre Model Development and Validation Study**

Raghavan Lavanya, MRCS(Edin),[1,4]* Yangqin Feng, PhD,[1,2]* Ten Cheer Quek, B.E.,[1] Quan V. Hoang, PhD,[1,3,5] Linda Yi-Chieh Poon, MD,[6] Jost B. Jonas, MD,[1,7] Ya Xing Wang, MD,[8,13] Vinay Nangia, FRCOphth,[9] Jin Wook Jeoung, PhD,[10] Sehie Park, MD,[10] SoYeon Kim, MD,[10] Benjamin Y Xu, PhD,[11] Sreenidhi Iyengar Munimadugu, MS,[11] Paul Mitchell, PhD,[12] Gerald Liew, PhD,[12] Yanin Suwan, MD,[13] Jirayu Hong-amata, [14] Sahil Thakur, PhD,[1] Monisha E Nongipur, PhD, [1,4] Tina Wong, PhD,[1,4] Rahat Husain, FRCOphth,[1] Ng Si Rui, FRCOphth,[1] Yamon Syn, FRCS,[1] Phey Feng Lo, FRCOphth,[1] Nicholas Tan Yi Qiang, FRCOphth,[1] Shaista Hussain, PhD,[2] Xiaofeng Lei, MSc,[2] Zhi Da Soh, PhD,[1] Marco Yu, PhD,[1] Haslina Hamzah, BSc,[15] Zizhou Wang, PhD,[2] Yan Wang, PhD,[16] Liangli Zhen, PhD,[2] Xinxing Xu, PhD,[2] Tien-Yin Wong, PhD,[1,8,17] Tin Aung, PhD,[1,3,4] Rachel S Chong, PhD,[1,4] #Yong Liu PhD,[2] #Ching-Yu Cheng, PhD.[1,3,4,5]

1 Singapore Eye Research Institute, Singapore National Eye Centre, Singapore
2 Institute of Advanced Intelligence and Computing, Agency for Science, Technology and Research, Singapore
3 Department of Ophthalmology, Yong Loo Lin School of Medicine, National University of Singapore, Singapore
4 Ophthalmology & Visual Sciences Academic Clinical Program (Eye ACP), Duke-NUS Medical School, Singapore
5 Centre for Innovation and Precision Eye Health, Yong Loo Lin School of Medicine National University of Singapore
6 Department of Ophthalmology, Kaohsiung Chang Gung Memorial Hospital, Kaohsiung, Taiwan
7 Rothschild Foundation Hospital, Paris, France
8 Beijing Visual Science and Translational Eye Research Institute (BERI), Beijing Tsinghua Changgung Hospital Eye Center, Tsinghua Medicine, Tsinghua University, Beijing, China.
9 Suraj Eye Institute, Nagpur, India

10 Department of Ophthalmology, Seoul National University Hospital, Soeul National University College of Medicine, Seoul, South Korea
11 USC Roski Eye Institute, Keck School of Medicine, United States of America
12 Centre for Vision Research, Westmead Institute for Medical Research, The University of Sydney, Sydney, New South Wales, Australia
13 Department of Ophthalmology, Ramathibodi Hospital, Mahidol University, Thailand
14 Faculty of Medicine Ramathibodi Hospital and Faculty of Engineering, Mahidol University, Thailand
15 SNEC Ocular Reading Centre, Singapore National Eye Centre, Singapore
16 School of Artificial Intelligence, Sichuan University, Chengdu, China
17 School of Clinical Medicine, Tsinghua Medicine, Tsinghua University, Beijing, China.

* Joint First Authors
# Joint Last Authors

**Conflict of Interest:**
Ching-Yu Cheng, Yong Liu and Xiaofeng Lei are cofounders of Eye.AI Pte Ltd. Tien-Yin Wong is a co-founder of EyRiS Pte Ltd. Benjamin Y. Xu has received consulting fees from Movu, Abbvie and Topcon. All other authors declare no competing interests.

**Corresponding author:**
Dr. Ching-Yu Cheng
Address: 20 College Road, The Academia, Level 6, Discovery Tower, Singapore 169856, Singapore.
Tel: +65 6576 7277, Fax: +65 6225 2568; Email: chingyu.cheng@duke-nus.edu.sg

**Abstract**

**Background**: Artificial intelligence (AI)–based systems for detecting glaucoma from colour fundus photographs (CFP) offer scalable and cost-effective tools for diagnosis and screening in both community and specialist settings. However, diagnostic performance often declines on external datasets due to inconsistent ground-truth definitions reflecting grader variability, population differences, and coexisting ocular conditions such as high myopia (HM). We developed and validated a Vision Transformer–based deep learning (DL) model to detect glaucoma from CFP across multi-ethnic cohorts with and without HM, while addressing these limitations.

**Methods:** A Vision Transformer (ViT-B/16) model incorporating predictive uncertainty estimation was developed using 56,483 CFPs (57·1% with myopia, 14·4% with HM). Ground-truth glaucoma labels were standardised using comprehensive clinical, imaging, and perimetry data to enhance diagnostic accuracy and generalisability. The model was extensively validated on sixteen independent datasets across three continents, including public, population-based, and hospital-based cohorts; four datasets included explicit HM labels.

**Findings:** The DL model achieved excellent internal performance with an area under the receiver operating characteristic curve (AUROC) of 98·7% (95% CI 98·2–99·1%), sensitivity 94·5%, and specificity 97·3%. External validation showed strong performance across sixteen datasets from eight countries with AUROCs ranging from 99·6% to 86·4%. In HM subgroups, internal performance remained robust (AUROC 97·8%, 95% CI 96·1–99·2%; sensitivity 94·8%; specificity 93·7%), with external validation across one population-based (Beijing Eye Study, AUROC 86·5%) and three hospital-based datasets from Taiwan, Thailand, and South Korea (AUROCs 93·3%, 91·8%, and 85·5%, respectively). In an exploratory clinical evaluation of HM eyes, the model demonstrated higher diagnostic accuracy than ophthalmologists and trained graders using CFP alone (accuracy 92·0% vs 70·0%; p =0·008) and achieved performance comparable to glaucoma specialists with access to full clinical, imaging, and perimetry results.

**Interpretation:** The DL model enables robust, generalisable glaucoma detection across myopic and non-myopic multi-ethnic populations, supporting AI-assisted screening in settings with high HM prevalence.

**Funding:** Singapore Ministry of Health's National Medical Research Council (OF-LCG MOH-001072–08)

## Research in context

### Evidence before this study:

Deep learning (DL) has shown promise in detecting glaucoma using colour fundus photographs (CFP), with supervised models achieving area under the receiver operating characteristic curve (AUROC) values of 0·94–0·99 in internal validation settings. However, their performance often declines significantly on external datasets, limiting the generalisability and scalability of artificial intelligence (AI) models for glaucoma screening. This sub-optimal performance is mainly due to variations in glaucoma definitions for ground truth, population characteristics, imaging protocols, and the presence of co-existing ocular diseases.

We conducted a comprehensive literature review of AI-based glaucoma detection studies using PubMed, Embase, and Google Scholar (publications up to August 2025). Search terms included "glaucoma detection," "deep learning," "glaucoma screening," "colour fundus photography," "high myopia," and "optic nerve head." Studies were included if they reported AI models trained on fundus images for glaucoma diagnosis, with or without inclusion of high myopia (HM).

Prior studies often relied on inconsistent ground truth definitions, often based solely on fundus photograph assessments by graders with varying levels of expertise, leading to diagnostic ambiguity. Furthermore, the presence of high myopia introduces additional complexity, as myopic optic discs can mimic glaucomatous features, contributing to increased false positive and false negative rates. With myopia rising globally, its impact on AI-based glaucoma detection remains a critical and unresolved challenge.

### Added value of this study

We present a Vision Transformer DL model (RetiGON) with a predictive uncertainty estimation, trained on a multi-ethnic, high-myopia-inclusive dataset (57·1% myopic, 14·4% HM images) with two key strengths. First, we employed standardized glaucoma definitions based on comprehensive clinical, imaging, and perimetry data, ensuring accurate and consistent labelling during training. This multi-modal approach for ground truth labels enhances diagnostic precision, improves reproducibility, and mitigates the ambiguity often seen in earlier AI studies that relied solely on fundus photographs for ground truth. Second, the inclusion of a high proportion of myopic (57·1%) and high myopic (14·4%) eyes makes this one of the first AI models specifically optimized for glaucoma detection in this high-risk subgroup. The model demonstrated strong generalizability across 16 external datasets from eight countries, including HM subgroups, and demonstrated higher diagnostic accuracy than ophthalmologists and trained graders in diagnosing glaucoma from CFP's in a sub-group of HM eyes (accuracy 92·0% vs 70·0%; p< 0·001). This study sets a new benchmark for AI-based glaucoma detection in populations with a high burden of myopia. Finally, the uncertainty-aware Vision Transformer classifier ensures that model decisions are accompanied by a quantitative measure of confidence, flags out unreliable predictions for human expert review, thereby improving reliability, facilitating human-AI collaboration, and reducing the risk of misclassification.

**Implications of all the available evidence:**

By integrating standardized glaucoma definitions and high-myopia training**,** this study represents a significant advancement in glaucoma screening for myopic populations**.** By integrating uncertainty estimation and enabling clinician review of ambiguous cases, the model aligns with real-world clinical decision-making, supporting the use of AI as a complementary tool to enhance clinical expertise in glaucoma care. Future research should focus on expanding training datasets to include more cases with advanced myopic degeneration across varied healthcare settings.

## INTRODUCTION

Glaucoma is the leading cause of irreversible blindness worldwide, affecting an estimated 111·8 million people by 2040.[1] Globally, more than half of all people with glaucoma remain undiagnosed, with rates rising up to 90% in the developing countries of Asia and Africa.[2] The disease often progresses silently without symptoms until its advanced stages, when central vision is affected. This diagnostic delay often leads to late-stage presentation with irreversible vision loss and a nearly fourfold increase in socioeconomic costs.[3] There is a pressing need for simple, cost-effective and scalable screening strategies to facilitate early diagnosis and reduce the burden of glaucoma-related blindness.

Among the diagnostic methods, colour fundus photography (CFP) is an inexpensive, and convenient tool for optic disc evaluation, making it ideal for low-cost, large scale glaucoma screening. Prior studies using conventional AI-based deep learning (DL) models for glaucoma detection have shown potential to improve diagnostic accuracy and minimize subjectivity, with supervised models achieving area under the receiver operating characteristic curve (AUROC) of 0·94–0·99 in internal validation.[4,5] However, their performance declines significantly on external datasets, dampening large scale initiatives to deploy these models for screening.[4,5]

The gap between research performance and real-world application stems from lack of a standardized definition of glaucoma, differences in population characteristics (ethnicity, disease prevalence, and data source), variations in image acquisition (camera type, field of view, area of interest (disc or macula centred), and image quality), and lower accuracy in eyes with co-existing ocular diseases.[5] For example, a key limitation of earlier studies has been their reliance on fundus photo-based ground truth labels assessed by graders with varying expertise, without confirmatory clinical, imaging, or visual field data.[4,5]

Secondly, the presence of co-existing ocular conditions, particularly myopia and high myopia (HM) add significant diagnostic ambiguity in glaucoma detection, often causing false positives (FP) and false negatives (FN).[6,7] Myopia is a major risk factor for primary open-angle glaucoma (POAG), with each 1-dioptre increase raising the risk of glaucoma by about 20%, with a steep, non-linear rise observed in HM.[8] Myopic eyes often exhibit characteristic structural alterations such as optic disc tilting, peripapillary atrophy, and thinner retinal nerve fibre layers (RNFL), which frequently overlap with features of a glaucomatous disc, making accurate diagnosis of glaucoma in myopic eyes challenging, especially in eyes with HM or with normal intraocular pressures (IOP).[9] Furthermore, Asia currently has the highest myopia rates, affecting up to 90% of young adults and 48% of older adults in countries like Singapore, Taiwan, South Korea, Japan and China.[10] By 2050, nearly 50% of the global population is projected to be myopic, with 10% having HM.[9] As myopia becomes a growing public health concern, it also presents a significant challenge for AI based-glaucoma detection.

To address the limitations of previous DL models, while accurately representing real-world populations with glaucoma and myopia, we developed and validated a vision transformer DL model for glaucoma detection using CFP, trained on a multi-ethnic dataset with both myopic and HM cases, using standardized glaucoma definitions based on comprehensive clinical, perimetry and imaging data.

## METHODS

### Datasets

The development datasets for our DL model ("RetiGON") included fundus images from the Singapore Epidemiology of Eye Diseases (SEED) Study and three clinical studies on glaucoma and HM from the Singapore National Eye Centre (SNEC). In brief, SEED is a well-established multi-ethnic, longitudinal, population-based study with 6 and 12-year follow-ups, representing the Malay, Indian, and Chinese populations ≥ 40 years of age in Singapore.[11] The three clinical studies from SNEC comprised prospective clinic-based studies on glaucoma, HM, and glaucoma in HM. Written informed consent was obtained from all participants in each study, and all studies adhered to the Declaration of Helsinki, with approval from the ethics committees. Permission to use the data was granted by the principal investigators.

The development dataset was curated to capture a broad clinical spectrum representative of the intended deployment setting. The SEED cohort provided a multi-ethnic population-based sample, while the SNEC clinical cohorts enriched the dataset with glaucoma and HM cases, ensuring sufficient representation of disease phenotypes and severity levels for robust model training.

Overall, the DL development dataset consists of 56,589 images from 16,364 multi-ethnic subjects, split into training (39,691 images), validation (5,524 images) and internal test sets (11,374 images) in a 70:10:20 ratio. Each subject contributed a maximum of 2 images (disc and macula-centred retinal photographs) per eye. All images from the same participant, including images from both eyes, were assigned exclusively to either the training, validation, or internal test set. During model training, each retinal photograph was used as an individual input image; however, participant-level separation ensured that correlated images from the same individual did not overlap across different datasets, thereby minimizing data leakage and overfitting.

**Diagnosis of glaucoma**

Glaucoma diagnosis was standardized across the SEED and SNEC studies, defined as the presence of characteristic structural abnormalities of the optic nerve head (characterized by neuroretinal rim thinning or notching and/or RNFL loss) with corresponding visual field (VF) defects, and in the absence of other explanation for the optic disc and VF appearances.[12] A glaucomatous VF defect was defined by an abnormal glaucoma hemifield test (GHT) and a cluster of ≥3 non-edge contiguous points with <5% probability in the pattern deviation plot on two separate occasions. A combination of slit lamp evaluation (Goldmann tonometry, gonioscopy, anterior segment examination), central corneal thickness (CCT) measurement (Ultrasound pachymetry, Advent; Mentor O & O Inc., Norwell, MA), ocular biometry (IOLMaster V3·01, Carl Zeiss Meditec AG, Jena, Germany), ocular imaging including fundus photo (Canon CR-1 Mark-II Nonmydriatic Digital Retinal Camera), optical coherence tomography (Cirrus HD-OCT, Carl Zeiss Meditec, Germany), and perimetry (Humphrey Visual Analyzer II's model 750, Carl Zeiss Meditec, Switzerland) were used to aid in the final diagnosis of glaucoma. When perimetric results were unreliable or unavailable, we defined glaucoma in accordance with the Category 2 of the International Society of Geographical and Epidemiological Ophthalmology (ISGEO) criteria;[13] (i.e. vertical cup-disc ratio (VCDR) > 99·5th percentile for our population[12]), with comprehensive specialist review of all clinical data and OCT imaging results and excluding other causes for the optic nerve damage. Glaucoma diagnoses were further corroborated by a review of clinical case notes from the Singapore National Eye Centre (SNEC), where available.

Prior to inclusion in the development dataset, all cases were meticulously reviewed by experienced glaucoma fellowship-trained ophthalmologists using all available clinical, imaging and perimetric data to ensure accurate ground truth labels. Borderline or doubtful cases, labelled as "glaucoma suspects," and ocular hypertension cases were excluded from the training dataset. The training dataset included images with co-existing ocular diseases (myopia, diabetic retinopathy (DR), age-related macular degeneration (AMD), epi-retinal membranes etc.) to enhance real-world applicability. High myopic subjects had spherical equivalent (SE) $\leq$ -6D and/or axial length (AL) $\geq$ 26 mm. We included cases with myopic macular degeneration (MMD) up to diffuse chorioretinal atrophy (Category 2 of the META-PM classification for pathologic myopia).[14]

**Validation datasets**

After developing and internally testing our RetiGON model, we further validated it using sixteen external, independent datasets from China, India, South Korea, Taiwan, Thailand, Spain, Australia and the United States of America. They included: *1)* three population-based studies: the Beijing Eye study (BES),[15] Central India Eye and Medical study (CIEMS),[16] and the Blue Mountains Eye Study (BMES);[17] *2)* nine publicly available datasets (REFUGE 2,[18] GAMMA,[19] LAG,[20] KEH,[21] Drishti-GS,[22] EyePACS,[23] PAPILA,[24] ODIR,[25] JSIEC [26]), and *3)* four hospital based datasets from Kaohsiung Chang Gung Memorial Hospital (KCGMH), Taiwan; Seoul National University Hospital (SNUH), South Korea; Ramathibodi Hospital (RH), Mahidol University, Thailand and USC Roski Eye Institute (REI), California, United States of America. The publicly available datasets from Asia (REFUGE 2, GAMMA, LAG, JSIEC) included myopic eyes (the exact numbers are unspecified), while the ODIR dataset specifically included 5% of eyes with pathological myopia. Four external datasets, namely KCGMH, SNUH, Ramathibodi Hospital, and BES included clear labels for eyes with HM, enabling model validation separately in HM eyes. **Table 1** summarizes the study participant demographics, geographic location, proportion of glaucoma cases, reference standard used for glaucoma diagnosis, fundus camera models, image resolution, and field of view for each external validation dataset. Additional demographic characteristics of the development and external validation datasets, including age, sex, and race/ethnicity where available, are provided in **Supplementary Table 1.** As these external datasets were derived from different studies with their own predefined protocols, the criteria used for glaucoma diagnosis varied across cohorts. The diagnostic criteria ranged from fundus photograph–based assessment, clinical optic disc assessment, combinations of fundus photographs with visual field testing and/or OCT, ISGEO classification, and other established dataset-specific reference standards.

External validation was performed using the original diagnostic reference standard defined by each contributing study.
The performance of our DL model was evaluated using the area under the receiver operating characteristic curve (AUROC) with 95% confidence intervals (CI), and sensitivity and specificity. The optimal prediction threshold was set by maximizing the G-Mean in this imbalanced dataset, balancing sensitivity (true positive (TP) rate) and specificity (true negative (TN) rate) to ensure clinically relevant performance. Positive predictive value (PPV) and negative predictive value (NPV) were computed as the proportion of correctly classified positive and negative cases, respectively, based on the model's confusion matrix (PPV = TP / [TP + FP]; NPV = TN / [TN + FN]).

**Design of the clinical study to compare the performance of our DL model to human experts in high myopic eyes**

A clinical study was conducted to compare model performance against human experts using 50 highly myopic eyes from an independent clinical study cohort recruited prospectively from a dedicated myopia glaucoma clinic. These eyes were not part of the model development, training, validation, or internal test datasets. Baseline demographics and ocular characteristics of the this HM cohort are provided in Supplementary Table 2.

Eight glaucoma specialists (three senior consultants, two mid-level consultants, and three fellows) and two experienced graders from the Ocular Reading Centre at SNEC (SORC) participated. In the first evaluation round (screening scenario), all human experts assessed the presence of glaucoma in HM eyes based solely on fundus photographs and were masked to the reference diagnosis, AI predictions, and all clinical information. The human experts performed their assessments independently, and no time limit was imposed for image interpretation. In the second round (diagnostic scenario), conducted two weeks later, the glaucoma specialists re-evaluated the same eyes but with access to additional clinical information, including IOP, CCT, AL, reliable VF and OCT scans. Throughout both evaluation rounds, glaucoma specialists remained masked to the reference diagnosis and AI predictions. The model's accuracy, sensitivity, and specificity were compared with those of human experts. Accuracy was defined as the percentage of correct predictions (sum of TP and TN cases) out of all assessments. Differences in diagnostic accuracy between RetiGON and human experts were assessed using a generalised linear mixed model (GLMM) to account for the paired and clustered nature of the grader study design.

### Development of the model

Our model employes a two-stage DL pipeline for robust glaucoma detection incorporating an uncertainty-aware classifier to enhance reliability and clinical applicability (**Figure 1).** It consists of *1)* a segmentation network (MobileNetV2-UNet)[27] for image quality control and optic disc localization, and *2)* a Vision Transformer (ViT)[28] backbone with an uncertainty-aware classifier that classifies the cropped optic disc image as glaucoma or non-glaucoma. The ViT-B/16 backbone was pretrained on ImageNet and fine-tuned on our development set, and a cost-sensitive loss up-weighting the minority glaucoma-positive class was used to address class imbalance. The uncertainty-aware component quantifies the confidence of each prediction using an entropy-based uncertainty score ($\mu$) derived from the model's output probability distribution. Cases with high predictive uncertainty (<10% of the dataset) are flagged as unreliable and referred for clinician review, thereby aligning the system with real-world clinical decision-making. The uncertainty-aware classifier ensures that model decisions are accompanied by a quantitative measure of confidence, improving reliability, facilitating human-AI collaboration, and reducing the risk of misclassification. To improve generalisation across the multi-ethnic, multi-device cohorts, the classifier was optimised with Stochastic Weight Averaging Densely (SWAD),[29] which seeks flat minima by densely averaging model weights and improved stability under external distribution shift. The details of the model development and ablation studies are elaborated in **Supplementary Appendix 1.** Performance comparison of different backbone architectures with SWAD but without the uncertainty module are provided in **Supplementary Table 3**. **Supplementary Table 4** demonstrates the ablation study of the ViT backbone model without SWAD and the uncertainty module across all datasets.

### Role of the funding source

The funding source had no role in study design, data collection, data analysis, data interpretation, or writing of the report.

## RESULTS

Of the 56,589 images in the development dataset, 106 were discarded during segmentation, leaving 56,483 images in the dataset. The mean (± SD) age of the participants was 60·3 ± 10·1 years, with 8,141(49·7%) male participants. It included 6,369 (38·9%) Chinese, 4,686 (28·6%) Malays, 5,299 (32·4%) Indians, and 10 (0·06%) from other ethnicities. There were 3,635 images with glaucoma and 52,848 images without glaucoma. Myopia was present in 32,256 images (57·1%), with 8,154 (14·4%) classified as HM. High myopia with glaucoma was seen in 1,648 (20·2%) of the images with HM. The mean (± SD**)** AL of the entire dataset was 23·8 ± 2·2 mm (range: 15·0–39·1 mm).

Prior to applying the uncertainty threshold, our RetiGON model achieved an AUROC of 97·8% (95% CI: 97·4–98·3%), with a sensitivity of 94·3%, specificity of 91·1%, positive predictive value (PPV) of 43·9 % and negative predictive value (NPV) of 99·6% in the internal test set. A small subset of highly uncertain predictions, identified using an entropy-based uncertainty score (<10% of the dataset), was excluded to mirror real-world deployment where such cases are referred for human review. After exclusion, the model's AUROC for detecting glaucoma was 98·7% (95% CI: 98·2–99·1), with a sensitivity of 94·5%, specificity of 97·3%, positive predictive value (PPV) of 71·3% and negative predictive value (NPV) of 99·6%.

The AUROC of RetiGON for glaucoma detection across external datasets ranged from 86·4% to 99·6% (**Table 2)**, with strong performance noted despite variations in cameras, resolution, and ground truth definitions. The highest AUROC was 99·6% (95% CI: 98·6–100·0%) in the GAMMA dataset and remained above 90·0% in most of the external datasets. The lowest performance was in the BMES dataset at 86·4% (95% CI: 81·7–90·8%). To further evaluate the impact of different reference standards on external validation performance, we stratified the external validation datasets according to the reference standard used for glaucoma diagnosis (**Supplementary Table 5**). For each reference standard category, predictions from all constituent datasets were pooled and the overall AUROC calculated. The highest pooled AUROC was observed in Group A (reference standard: comprehensive clinical assessment; 95·2%), followed by Group B (reference standard: clinical assessment with HVF and/or OCT; 93·8%) and Group C (reference standard: fundus photograph–based assessment; 93·1%). The pooled AUROC was significantly higher in Group A than in Group B (p=0·009) and Group C (p<0·0001), and was also higher in Group B than in Group C (p=0·01).The uncertainty-aware filtering approach was consistently applied across all external validation datasets, leading to marginal but consistent improvements in performance (**Supplementary Table 6**). Calibration analysis demonstrated good agreement between predicted probabilities and observed outcomes, with an Expected Calibration Error (ECE) of 0·033 and Brier score of 0·026 in the internal test dataset (**Supplementary Figure 1** and **Supplementary Table 7**). Temperature scaling resulted in minimal change in calibration (ECE 0·033 to 0·032; optimal temperature 0·979).

**Analysis of uncertain predictions**

Analysis of factors associated with uncertainty flagging demonstrated that high myopia was associated with an increased likelihood of being flagged in the internal dataset (OR 1·56, 95% CI 1·31–1·86), with a similar trend observed in external datasets with available high-myopia labels. After adjustment, high myopia remained independently associated with uncertainty flagging, while the effects of camera type, age, sex, and ethnicity were comparatively modest (**Supplementary Figure 2A and 2B**).

To further characterise cases identified by the uncertainty module, we analysed the most uncertain false-positive and false-negative predictions (**Supplementary Figure 3 and 4**). Highly uncertain cases were frequently associated with challenging optic disc features and image-related factors, including large physiological cupping, peripapillary atrophy, high myopia-related disc changes, atypical optic disc appearance, off-centre image acquisition, and imaging artefacts.

**Performance in highly myopic eyes**

In the sub-group of HM eyes in the internal test set, our algorithm showed excellent performance with an AUROC of 97·8% (95% CI: 96·1–99·2%) with a sensitivity of 94·8%, and a specificity of 93·7%. Strong external validation performance was observed in the KCGMH (Taiwan) and Ramathibodi Hospital (Thailand) datasets, with AUROC of 93·3% (95% CI: 89·8–96·5) and 91·8% (95% CI: 82·9–97·9%) respectively, supporting robust generalisability in HM populations (**Table 3**). Good to fair performance was noted in the BES and SNUH datasets (AUROC of 86·5% and 85·5% respectively). **Figure 2** (A-B) and **Figure 3** (A-B) shows saliency maps from the non-HM and HM cases, highlighting areas influencing the model's predictions. Regions with greater influence appear in red and those with lesser impact in blue, with correct focus on glaucomatous rim loss or an enlarged optic cup consistent with clinical patterns.

**Model performance compared to human experts in highly myopic eyes**

In the first evaluation round using only CFPs (screening scenario), RetiGON achieved an AUROC of 96·3% (95% CI: 91·7–100·0%), with accuracy, sensitivity, and specificity of 92·0% each for glaucoma detection. After excluding five eyes with high uncertainty scores, the AUROC increased to 97·8%. In contrast, the two professional graders at SORC achieved sensitivities of 58·3% and 65·2% and specificities of 80·0% and 84·0%, while glaucoma specialists demonstrated wider variability, with sensitivities from 55·0% to 84·0% and specificities from 36·0% to 96·0% (**Figure 4A**).

Even when retaining uncertain cases, RetiGON achieved higher diagnostic accuracy than the human experts (p = 0·008), achieving an accuracy of 92·0% compared with a mean accuracy of 70·6% (range 57·4–81·2%) among glaucoma specialists and trained graders, using CFP alone. When limited to senior and mid-level glaucoma consultants, the mean accuracy was 71·0%, and RetiGON remained significantly superior (p= 0·01).

In the second evaluation round (diagnostic scenario), when additional clinical, imaging and perimetry information was provided, the glaucoma specialists' performance improved markedly, with mean accuracy increasing to 84·2% (range 64·0–92·0%), sensitivity ranging from 76·0% to 96·0%, and specificity from 32·0% to 100·0% (**Figure 4B**). Under these conditions, there was no statistically significant difference between RetiGON and the glaucoma specialists (p = 0·064), highlighting that contextual clinical data helps bridge the remaining gap between algorithmic and human expertise.

**Misclassified images in the internal dataset**

Further analysis of the internal dataset was conducted to evaluate RetiGON's performance and identify the underlying causes of false positives (FP) and false negatives (FN) (**Table 4**). The most common cause for a FP result was a large physiological cup (45·9%), followed by HM eyes with extensive peripapillary atrophy (29·2%) and other eye conditions, including diabetic retinopathy with large laser scars close to disc, congenital optic disc abnormalities, non-glaucomatous optic atrophy, and retinal diseases (13·6%). Additionally, 11·3% of FP cases had a normal fundus misclassified as glaucoma.

False negatives were most frequently seen when glaucomatous optic neuropathy coexisted with other eye diseases (48·6%), such as diabetic retinopathy  and age-related macular degeneration (27·0%), and HM eyes with tilted discs (21·6%). Another key FN pattern occurred in eyes with very small discs, where the perceived small cup-to-disc (CD) ratio underestimated glaucomatous changes, leading to misclassification (29·8%). The model also misclassified cases with subtle RNFL defects in infero- or supero-temporal sectors without clear cupping or rim notching (16·2%) and in two cases (5·4%) with disc haemorrhages.

**Misclassified images in the SNUH High Myopia dataset**

Among the 77 highly myopic (HM) eyes in the SNUH cohort, RetiGON correctly classified 57 (74·0%) eyes and misclassified 20 (26·0%) eyes. All misclassified eyes were false negatives and predominantly represented early-stage glaucoma, accounting for 14 of 20 eyes (70·0%) (**Supplementary Table 8**). Compared with correctly classified eyes, misclassified eyes had a significantly smaller vertical cup-to-disc ratio (VCDR; 0·60 ± 0·16 vs 0·71 ± 0·15; p=0·012). They also had better visual field mean deviation (MD; −4·31 ± 3·79 dB vs −6·04 ± 6·71 dB; p=0·277) and thicker retinal nerve fibre layer (RNFL; 74·5 ± 10·09 µm vs 73·96 ± 13·95 µm; p=0·875), although these differences were not statistically significant. Representative examples of these challenging cases are shown in **Supplementary Figure 5**.

**DISCUSSION**

We developed and validated a DL model, RetiGON, for detecting glaucoma from CFP, using a population-representative, multi-ethnic dataset with two key strengths. First, it employed standardized glaucoma definitions based on comprehensive clinical, imaging, and perimetry data, ensuring accurate and consistent labelling during training. This  approach enhances diagnostic precision, improves reproducibility, and mitigates the ambiguity often seen in earlier AI studies that relied solely on fundus photographs for ground truth labels. Second, the inclusion of a high proportion of myopic (57·1%) and HM (14·4%) eyes in the training dataset makes this one of the first AI models specifically optimized for glaucoma detection in this high-risk subgroup. Our model achieved higher diagnostic accuracy than expert graders in an exploratory clinical evaluation of HM patients, underscoring its potential to address diagnostic challenges unique to myopic eyes.

We adopted a stringent binary classification approach for glaucoma detection to identify high-risk individuals needing timely intervention, as undiagnosed glaucoma remains a significant public health concern. Studies in Singapore indicate that 72% of glaucoma cases go undetected, with rates exceeding 90% in the Malay community.[30] Of those with undiagnosed glaucoma, 56% had moderate or worse VF loss in at least one eye, and 4·1% were already blind in one eye due to end-stage disease. Secondly, a recent evaluation showed that opportunistic glaucoma screening via CFP by trained graders in Singapore's Integrated Diabetic Retinopathy Programme (SiDRP) resulted in high FP rates and excessive unnecessary referrals, with sensitivity of 81·6%, specificity of 50·6%, and a PPV of just 14·0%.[31] We prioritized high specificity to minimize false-positive referrals, which can burden healthcare systems and undermine trust in screening programs. By excluding glaucoma suspects and borderline cases, the focus remains on confirmed glaucoma cases, enabling timely intervention. Although this approach may reduce sensitivity, it enhances overall screening efficiency by detecting more true glaucoma cases than current methods. This approach balances the need for efficient resource use with effective case detection.

Accurate glaucoma diagnosis in myopic eyes remains a significant challenge due to overlapping structural features that obscure disease detection. Additionally, VF defects may mimic glaucomatous patterns, while standard imaging modalities such as OCT are less reliable due to confounding factors like staphyloma, MMD, and lack of normative data for these anatomically distinct eyes. Our model, trained on 57·1% myopic and 14·4% HM cases, demonstrated strong performance, achieving an AUROC of 97·8% internally and maintaining high accuracy in external datasets. While AUROC values were slightly lower in HM cases compared to non-HM cases, they remained robust with AUROC of 93·3% in the KCGMH, 91·8% in Ramathibodi Hospital, 86·5% in BES dataset and 85·5% in SNUH dataset, ensuring reliability across hospital and population-based studies. Additionally, it performed well in public datasets with myopic eyes like GAMMA (50% myopic cohort), REFUGE-2, JSIEC and LAG, handling challenging cases like pathological myopia in the ODIR dataset and even some MMD category 3 cases in KCGMH despite not being part of training. Detailed error analysis of the SNUH highly myopic (HM) cohort demonstrated that misclassified glaucomatous HM eyes predominantly represented early-stage glaucoma. These findings highlight the inherent challenges of detecting early glaucomatous changes in highly myopic eyes, where myopia-related structural alterations of the optic nerve head can obscure the subtle features of early glaucoma. Such borderline cases remain challenging even for experienced glaucoma specialists and often require longitudinal follow-up with serial structural and functional assessments before a definitive diagnosis can be established.

Beyond improving diagnostic performance, the incorporation of uncertainty estimation provides an additional safety mechanism for AI-assisted glaucoma screening by identifying cases where automated predictions may require clinician review. Calibration analysis demonstrated good agreement between predicted probabilities and observed outcomes, supporting the use of entropy derived from these probabilities as a measure of predictive uncertainty. Further analysis showed that uncertainty flagging was more frequent in highly myopic eyes, consistent with the recognised diagnostic challenges associated with myopic optic disc changes. Qualitative review of uncertainty-flagged cases demonstrated that these cases frequently represented clinically challenging scenarios, including large physiological cupping, highly myopic eyes with disc tilt and extensive peripapillary atrophy, subtle glaucomatous features, and reduced image quality. These findings support an uncertainty-guided workflow in which challenging cases are referred for human review rather than relying solely on automated predictions. The clinical evaluation comparing the performance of RetiGON with human experts showed that RetiGON achieved high diagnostic accuracy in HM eyes, underscoring its reliability in challenging clinical scenarios. While RetiGON outperformed clinicians using images alone, ophthalmologists achieved comparable accuracy when provided with full clinical context, indicating that algorithmic precision and human judgment are distinct yet complementary components of diagnostic reasoning. Together, these results highlight a collaborative rather than competitive relationship between AI and clinician expertise in glaucoma detection.

Despite several challenges, our RetiGON model demonstrated adaptability, emphasizing the need for HM-inclusive training and setting a precedent for AI models to address anatomical variations. Previous studies had limited HM representation, for instance, Shibata et al. included only 55 HM cases,[32] while Kim et al. relied on OCT and demographic data (age, sex, AL and mean deviation (MD) values from visual fields) for detection.[33] More recently, Chiang et al. trained a CNN model on 3,088 HM images (AUROC 89%, sensitivity 81·0%, specificity 83·2%) but lacked external validation and excluded all cases of pathological myopia, limiting generalizability.[34] In contrast, our model sets a new benchmark for AI-based HM glaucoma detection. Saliency mapping confirmed its ability to focus on areas of the optic cup even in eyes with tilted discs, enhancing interpretability and diagnostic confidence.

Limitations of our study include that the development cohort consisted predominantly of Singaporean participants (Chinese, Malay, and Indian), and the majority of external validation datasets were derived from Asian populations. Nevertheless, external validation also included four non-Asian cohorts: the ethnically diverse EyePACS dataset, which comprised predominantly Latin American participants together with White, African, Asian, and other ethnic groups; a clinical cohort from the USC Roski Eye Institute in the United States; and the predominantly Caucasian BMES and PAPILA cohorts. While performance was comparatively lower in BMES and PAPILA, these differences are likely multifactorial and may reflect variations in disease spectrum, image acquisition protocols, image quality, case mix, and reference standards, in addition to population characteristics. Therefore, further validation in larger and geographically diverse non-Asian populations remains important before widespread clinical deployment.

Second, although our development dataset used a standardized multimodal reference standard, the external validation datasets were derived from different studies and employed different diagnostic reference standards according to their original study protocols. These differences may influence direct comparisons of performance metrics across datasets and should be considered when interpreting model performance across cohorts. However, evaluation across datasets with diverse reference standards provides insight into model performance across independently curated cohorts and different clinical and research settings.

Third, although RetiGON demonstrated robust performance in highly myopic eyes overall, performance was reduced in the SNUH high-myopia cohort, which contained a greater proportion of subtle early glaucoma cases. This highlights the ongoing challenge of detecting early glaucomatous changes in highly myopic eyes. Fourth, the exploratory comparison of RetiGON with glaucoma specialists and trained graders was conducted in a relatively small sample size cohort, limiting the generalisability of these findings. Larger prospective clinical studies are required to further evaluate model performance relative to clinicians in diverse clinical settings. Finally, advanced myopic maculopathy (META-PM grades 3 and 4) was not represented in the development dataset. Future studies should evaluate model performance in eyes with more advanced myopic degeneration.

In conclusion, this study demonstrates a robust DL model for glaucoma detection with high diagnostic accuracy across diverse populations and challenging HM subgroups, marking a significant advancement in reliable glaucoma screening, particularly in high myopia-prevalent regions. By integrating uncertainty awareness and enabling human review of ambiguous cases, this model aligns with real-world clinical decision-making and supports the use of AI as a complementary tool that enhances, rather than replaces clinician expertise in glaucoma care. Future work should expand training datasets to include more anatomically heterogenous HM cases with advanced myopic degeneration.

**Data sharing statement**

De-identified participant data used for model development may be made available by the corresponding author upon reasonable request, subject to institutional approval and applicable data-sharing agreements. As the RetiGON algorithm has been licensed commercially, the code and related source materials are not publicly available. For academic research purpose, controlled access to the algorithm through a docker-based environment may be provided upon reasonable request, subject to institutional approvals. The external datasets used in this study include publicly available datasets accessible through their respective repositories, as well as non-public population-based and hospital-based datasets accessible only through their respective data custodians and subject to institutional and data-use agreements. Relevant references for these datasets are provided in the manuscript.

**Author contribution statement**

C-YC, LY and TA conceptualized the study. RL, YF, XX, RSC, LY and C-YC designed the study. RL, YF, TCQ, QVH, LY-CP, JBJ, YXW, VN, JWJ, SP, SYK, BYX, SIM, PM, GL, YS, JH-A, ST, MEN, TW, RH, NSR, YS, PFL, NTYQ, SH, XL, ZDS, HH, TA, RSC, C-YC collected the data. YF, SH, XL, ZW, YW, LZ, XX, and YL developed the algorithm. RL, YF, TCQ, SH, XL, MY analysed the data. RL, YF, SH, TYW, RSC and C-YC drafted the manuscript. RL, TCQ, ST, ZDS, C-YC confirm they had access to the raw datasets of the SEED study. RL, TCQ, QVH, ST, ZDS, RSC confirm they had access to the raw datasets of clinical studies on glaucoma and high myopia from SNEC. YXW and JBJ confirm they had access to the raw datasets of the BES. VN and JBJ confirm they had access to the raw datasets of the CIEMS. PM and GL confirm they had access to BMES-4 dataset. LY-CP confirms she had access to the raw datasets of KCGMH, Taiwan. JWJ, SP, SYK confirm they had access to the raw data sets of the SNUH, South Korea. BYX and SIM confirm they had access to the raw data sets of the USC-REI, California. YS and JH-A confirm they had access to the raw data sets of the RHMU, Thailand. RL, YF, TCQ, SH, XL, XX, YL, and C-YC accessed and verified each dataset during the course of the study. All authors approved the final manuscript.

**Declaration of interest**

Ching-Yu Cheng, Yong Liu and Xiaofeng Lei are cofounders of Eye.AI Pte Ltd. Tien-Yin Wong is a co-founder of EyRiS Pte Ltd. Benjamin Y. Xu has received consulting fees from Movu, Abbvie and Topcon. All other authors declare no competing interests.

**Acknowledgement**

This study was funded by Singapore Ministry of Health's National Medical Research Council (OF-LCG MOH-001072–08)

## TABLES

**Table 1**: Characteristics of the different datasets used in development and validation

| Dataset | Country | Fundus camera | Field of View and/or Resolution in pixels | Number of images | Proportion of glaucoma cases in the dataset (%) | Reference standard used for Glaucoma diagnosis | Classification of Glaucoma |
|---|---|---|---|---|---|---|---|
| Development dataset for our AI model (RetiGON) | Singapore | Canon CR-DGi 10D, Topcon TRC NW8, Triton | 30, 45 degrees<br>3072 × 2048<br>3888 × 2592<br>3504 × 2336 | 56,483 | 6·4 | Clinical, VF, OCT, fundus photos with longitudinal follow-up data | G, NG |
| **Population based studies** | | | | | | | |
| BES—2[15] | China | Canon CR6-45NM | 45 degrees<br>3888 × 2592 | 10,724 | 3·6 | ISGEO classification | G, NG |
| CIEMS[16] | India | Zeiss FF450 | 50, 20 degrees<br>1280 × 1024 | 12,040 | 1·4 | Optic disc assessment on fundus photos | G, NG |
| BMES—4[17] | Australia | Canon CF-60 DSi | 40 degrees<br>4243 × 3294 | 3,518 | 2·6 | Clinical optic disc assessment, VF | G,NG |
| **Publicly available datasets** | | | | | | | |
| GAMMA[19] | China | Kowa,<br>Topcon TRC NW400 | 2000 × 2992<br>1934 × 1956 | 100 | 50·0 | Fundus photo, IOP, VF, OCT | G, NG |
| REFUGE—2[18] | China | Visucam, Canon CR-2, Kowa, Topcon | 45 degrees | 1,005 | 10·4 | Fundus photo, IOP, VF, OCT | G, NG |
| KEH[21] | South Korea | Nidek AFC-330 | 800 × 800 | 1,544 | 48·9 | Clinical, HVF and/or OCT or red free RNFL photography | G, NG |
| DRISHTI—GS[22] | India | NA | 30 degrees 2047 × 1760 | 101 | 69·3 | Fundus photos | G, NG |
| LAG[20] | China | Topcon, Canon, Zeiss | 500 × 500 | 4,439 | 31·0 | Clinical optic disc assessment, IOP, VF | G, NG |
| EyePACS[23] | USA | Variety of cameras | 45 degrees | 100,109 | 3·1 | Ten structural features on fundus photos | G, NG |
| PAPILA[24] | Spain | Topcon TRC-NW400 | 30 degrees 2576 × 1934 | 425 | 20·8 | Clinical data | G, NG |
| ODIR[25] | China | Canon, Zeiss, Kowa | Variable | 5,599 | 6·3 | Fundus photos | G, NG |
| JSIEC[26] | China | Zeiss FF450 plus IR, Topcon TRC-50DX | 35–50 degrees | 997 | 1·3 | Fundus photos | GS, NG |
| **Hospital based datasets** | | | | | | | |
| KCGMH | Taiwan | Topcon TRC 50DX | 50 degrees | 298 | 50·0 | Clinical, VF, OCT, fundus photos | G, HMG, NG, HMNG |
| SNUH | South Korea | Topcon, Kowa VX10-A, Zeiss Visucam | 45 degrees | 197 | 50·8 | Clinical, VF, OCT, fundus photos | G, HMG, NG, HMNG |
| USC REI | USA | Zeiss FF450 plus IR, Megavision 50DX/50 with NV Fundus | 20, 50 degrees | 494 | 50·0 | Clinical, VF, OCT, fundus photos | G,NG |
| RH | Thailand | Eidon, Kowa | 30, 50 degrees | 575 | 63·3 | Clinical, VF, OCT, fundus photos | G, HMG, NG, HMNG |

G: Glaucoma; NG: Glaucoma; GS: Glaucoma suspects
HM: High Myopia; HMG: High Myopia with Glaucoma; HMNG: High Myopia No (without) Glaucoma
VF: Visual Fields; OCT: Optical Coherence Tomography; NA: Not available
ISGEO: International Society of Geographical and Epidemiological Ophthalmology
BES—2: Beijing Eye Study (10 year follow-up); CIEMS: Central India Eye and Medical Study; BMES—4: Blue Mountain Eye Study (15 year follow-up)
KCGMH: Kaohsiung Chang Gung Memorial Hospital; SNUH: Seoul National University Hospital; USC REI: USC Roski Eye Institute; RH: Ramathibodi Hospital, Thailand

**Table 2:** Performance of the model in automated glaucoma detection in all datasets

| Dataset | AUROC (%) | 95% Confidence Intervals (%) | Sensitivity (%) | Specificity (%) |
|---|---|---|---|---|
| Internal test dataset | 98·7 | 98·2–99·1 | 94·5 | 97·3 |
| **Population based datasets** | | | | |
| BES—2 | 92·1 | 90·7–93·5 | 87·4 | 84·8 |
| CIEMS | 89·9 | 87·2–92·6 | 78·3 | 87·3 |
| BMES—4 | 86·4 | 81·7–90·8 | 71·1 | 87·8 |
| **Publicly available datasets** | | | | |
| GAMMA | 99·6 | 98·6–100·0 | 95·5 | 97·8 |
| REFUGE—2 | 95·3 | 92·0–98·1 | 85·1 | 97·2 |
| KEH | 91·6 | 90·1–93·0 | 86·5 | 82·2 |
| DRISHTI—GS | 90·0 | 81·6–97·1 | 87·5 | 88·9 |
| LAG | 95·0 | 94·3–95·7 | 86·0 | 93·3 |
| EyePACS | 93·3 | 93·0–93·7 | 91·4 | 88·6 |
| PAPILA | 88·4 | 84·0–92·7 | 76·0 | 84·4 |
| ODIR | 90·2 | 88·3–91·8 | 85·5 | 87·4 |
| JSIEC | 98·3 | 97·1–99·3 | 100·0 | 96·3 |
| **Hospital based datasets** | | | | |
| KCGMH | 95·9 | 93·7–97·7 | 89·1 | 90·8 |
| SNUH | 95·5 | 92·7- 98·3 | 88·2 | 93·2 |
| USC REI | 90·0 | 87·0–93·2 | 83·8 | 85·6 |
| RH | 88·2 | 84·8–91·4 | 76·3 | 91·9 |

AUROC: Area under the receiver operating characteristic curve
BES—2: Beijing Eye Study; CIEMS: Central India Eye and Medical Study; BMES—4: Blue Mountain Eye Study
KCGMH: Kaohsiung Chang Gung Memorial Hospital; SNUH: Seoul National University Hospital; USC REI: USC Roski Eye Institute; RH: Ramathibodi Hospital, Thailand

**Table 3:** Performance of the deep learning model ( RetiGON) in highly myopic eyes

| Dataset | AUROC (%) | 95% Confidence Intervals (%) | Sensitivity (%) | Specificity (%) |
|---|---|---|---|---|
| Internal test dataset (All) | 98·7 | 98·2–99·1 | 94·5 | 97·3 |
| Internal test dataset (HM) | 97·8 | 96·1–99·2 | 94·8 | 93·7 |
| Internal test dataset (Non-HM) | 98·7 | 98·2–99·2 | 94·1 | 97·8 |
| **External datasets** | | | | |
| KCGMH (All) | 95·9 | 93·7–97·7 | 89·1 | 90·8 |
| KCGMH (HM) | 93·3 | 89·8–96·5 | 85·7 | 87·4 |
| KCGMH (Non-HM) | 99·2 | 97·3–100·0 | 95·6 | 97·8 |
| BES—2 (All) | 92·1 | 90·7–93·5 | 87·4 | 84·8 |
| BES—2 (HM) | 86·5 | 74·7–96·4 | 73·3 | 84·1 |
| BES—2 (Non-HM) | 92·4 | 90·8–93·8 | 86·1 | 85·5 |
| SNUH (All) | 95·5 | 92·7–98·3 | 88·2 | 93·2 |
| SNUH (HM) | 85·5 | 76·8–94·2 | 66·0 | 100 |
| SNUH (Non-HM) | 97·8 | 95·2–99·9 | 92·5 | 98·5 |
| RH (All) | 88·2 | 84·8–91·4 | 76·3 | 91·9 |
| RH (HM) | 91·8 | 82·9–97·9 | 90·9 | 82·6 |
| RH (Non-HM) | 87·9 | 84·5–91·0 | 76·0 | 92·5 |

AUROC: Area under the receiver operating characteristic curve
HM: High Myopia; Non-HM: Non-High Myopia; All: Entire dataset
KCGMH: Kaohsiung Chang Gung Memorial Hospital, Taiwan; BES—2: Beijing Eye Study; SNUH dataset: Seoul National University Hospital, South Korea; RH: Ramathibodi Hospital, Thailand

**Table 4:** Reasons for misclassification by the deep learning model ( RetiGON)

| Misclassification Type | Cause/Pattern | Frequency | Percentage |
|---|---|---|---|
| **False Positives (FP)** | Physiologically large cupping | 118 | 45·9% |
| | High myopia with extensive peripapillary atrophy | 75 | 29·2% |
| | Other eye conditions (e.g., diabetic retinopathy, congenital optic disc abnormalities, non-glaucomatous optic atrophy, retinal diseases) | 35 | 13·6% |
| | Normal fundus | 29 | 11·3% |
| **Total FP cases** | | 257 | 100·0% |
| **False Negatives (FN)** | GON co-existing with other eye diseases | 18 | 48·6% |
| | - Diabetic retinopathy and AMD | 10 | 27·0% |
| | - High myopia with tilted discs | 08 | 21·6% |
| | Small discs and perceived small cup sizes, leading to underestimated glaucomatous changes | 11 | 29·8% |
| | Subtle RNFL defects in infero-and/or supero-temporal sectors without corresponding glaucomatous optic nerve damage | 06 | 16·2% |
| | Disc haemorrhages | 02 | 5·4% |
| **Total FN Cases** | | 37 | 100·0% |

GON: Glaucomatous optic neuropathy
AMD: Age-related macular degeneration
RNFL: Retinal nerve fibre layer

**Figures Title and Legend**

**Figure 1:** Architecture and workflow of the deep learning model ( RetiGON) for glaucoma detection

**Legend for Figure 1:** The deep-learning system comprises: (Stage 1) Fundus photographs from multiple datasets for training (A–D) processed using a MobileNetV2-UNet segmentation network for image quality control and optic disc localisation; (Stage 2) Robust classification of glaucoma versus non-glaucoma images using a Vision Transformer with uncertainty-based prediction. Model weight optimisation performed using stochastic weight averaging densely (SWAD) to enhance generalisability; (Stage 3) During clinical deployment at hospital or primary care sites (E), model outputs glaucoma (G) or non-glaucoma (NG) predictions along with an entropy-based uncertainty score (μ). Predictions with uncertainty exceeding a defined threshold (θ) are flagged as unreliable and referred for clinician review.

**Figure 2(A-B):** Segmented fundus photo and corresponding saliency map in non–high myopic eyes showing regions prioritized by the model (neuroretinal rim) for glaucoma detection.
**Legend:** High-impact regions influencing the prediction are shown in red, and lower-impact areas in blue.

**Figure 3 (A-B):** Segmented fundus photo and corresponding saliency maps in highly myopic eyes, highlighting the model's focus on the optic disc cup for glaucoma detection

**Legend**: High-impact regions influencing the prediction are shown in red, and lower-impact areas in blue.

**Figure 4 (A-B):** Comparison of performance of fundus photo–based deep learning model (RetiGON) and human experts for glaucoma detection in high myopia
**Legend:** (A) Area under the receiver operating characteristic (AUROC) curve comparing the performance of fundus photo–based deep learning model (RetiGON) to glaucoma specialists and trained graders with access to fundus photographs only. (B) AUROC curve comparing the same fundus photo–based model (RetiGON) to glaucoma specialists with access now to complete clinical data including perimetry and imaging results). The deep learning model achieved an AUROC of 0·963, with the optimal Geometric Mean threshold (0·857) indicated on the curve. Coloured points represent the sensitivity and specificity of individual human experts, including glaucoma specialists (senior consultants, consultants, and glaucoma fellows) and trained graders.